\documentclass[10pt,twocolumn,letterpaper]{article}

\usepackage[pagenumbers]{cvpr} 

\usepackage{graphicx}
\usepackage{amsmath}
\usepackage{amssymb}
\usepackage{booktabs}
\usepackage{multirow} 
\usepackage{xcolor, colortbl}
\definecolor{almond}{rgb}{0.94, 0.87, 0.8}
\usepackage{pifont} 
\usepackage{float}

\usepackage[pagebackref,breaklinks,colorlinks]{hyperref}

\usepackage[capitalize]{cleveref}
\crefname{section}{Sec.}{Secs.}
\Crefname{section}{Section}{Sections}
\Crefname{table}{Table}{Tables}
\crefname{table}{Tab.}{Tabs.}

\def\cvprPaperID{11273} 
\def\confName{CVPR}
\def\confYear{2022}

\begin{document}

\title{Efficient Self-Ensemble for Semantic Segmentation}

\author{Walid Bousselham \\ 
\and Guillaume Thibault \\ 
\and Lucas Pagano \\ 
\and Archana Machireddy \\ 
\and Joe Gray \\ 
\and Young Hwan Chang \\ \\
Oregon Health and Science University\\
Portland, OR, USA\\
\and Xubo Song\\
}

\maketitle


\begin{abstract}
 Ensemble of predictions is known to perform better than individual predictions taken separately. However, for tasks that require heavy computational resources, \textit{e.g.} semantic segmentation, creating an ensemble of learners that needs to be trained separately is hardly tractable. In this work, we propose to leverage the performance boost offered by ensemble methods to enhance the semantic segmentation, while avoiding the traditional heavy training cost of the ensemble. Our self-ensemble approach takes advantage of the multi-scale features set produced by feature pyramid network methods to feed independent decoders, thus creating an ensemble within a single model. Similar to the ensemble, the final prediction is the aggregation of the prediction made by each learner. In contrast to previous works, our model can be trained end-to-end, alleviating the traditional cumbersome multi-stage training of ensembles. Our self-ensemble approach outperforms the current state-of-the-art on the benchmark datasets  Pascal Context and COCO-Stuff-10K for semantic segmentation and is competitive on ADE20K and Cityscapes. Code is publicly available at \href{https://github.com/WalBouss/SenFormer}{github.com/WalBouss/SenFormer}.
\end{abstract}

\section{Introduction}
\label{sec:intro}

Semantic segmentation is the task of assigning each pixel of an image with a semantic category, and therefore is close to the task of image classification. Its many applications include robotics, autonomous cars, medical application, augmented reality and more. Most segmentation methods follow an Encoder-Decoder scheme. The encoder extracts the relevant features of the image to characterize each pixel, a process usually involving down-sampling the feature maps to increase the receptive field of the model. The decoder up-samples the feature maps to both recover the spatial information and produce a per-pixel classification. In \cite{FCN14} the authors extended this procedure to fully convolutional network (FCN), which paved the way for later work to achieve impressive results on various segmentation datasets and has since dominated the field of semantic segmentation, let it be for medical \cite{ronneberger2015u,Hesamian2019DeepLT}, self-driving cars \cite{siam2017deep} or robotics applications \cite{hong2018virtualtoreal}. Follow up work mainly focused on enhancing FCN to mitigate the inherent locality of the convolution operation. Some examples are the atrous convolution that introduces holes in convolution kernel \cite{chen2017deeplab,chen2017rethinking}, the pyramid pooling module (PPM) that aggregates context information using different kernel pooling layers, and \cite{uper2018} that combine the PPM and the feature pyramid network (FPN) \cite{lin2017feature} to capture context information at different resolutions.

The starting observation of this paper was that the combined use of a backbone and an FPN-like method \cite{lin2017feature,liu2018path,ghiasi2019fpn,tan2020efficientdet,wang2020SEPC,huang2021fapn} allows extracting multiple features sets at different scales for a single image with a unique forward pass. Furthermore, in \cite{lin2017feature} the authors show that these features are both semantically and spatially strong at each level of the feature pyramid. Consequently, one has access to multiple features representations of the same image that carry different contextual information at different scale and are loosely correlated (as shown in\cite{wang2020SEPC}). This raises questions about the optimal way to use these multi-scales features. A canonical use is UperNet \cite{uper2018}, which concatenates the multi-scale feature maps before feeding them to a decoder. However, this paper argues that the "features fusing" strategy consisting of merging the different sets of features maps and letting the model decide which one is important is sub-optimal and often computationally expensive. Indeed, in UperNet, the four pyramid levels are concatenated and merged by a convolution, which by its own involves 155G FLOPs, thus making the "features fusing" strategy FLOPs intensive. Moreover, we hypothesize that a single decoder cannot fully take advantage of the multi-scale features that contain different views of the same objects of interest. Hence, the model may focus on one view and overlook valuable features. This "multi-view" hypothesis is indeed supported by a recent study: in \cite{allenzhu2021understanding} the authors argue that in vision datasets, objects can be recognized using multiple views and show that in the context of image classification, for a given weight initialization, a model will learn to focus on particular views while discarding others.



To overcome the limitations of "features fusion" strategies, we propose and study a different approach to exploit the FPN multi-scale features. Our approach feeds independent decoders with features coming from different levels of the feature pyramid, and then combine the segmentation maps together, hence avoiding expensive features fusion operations. Since the inputs to the learners (i.e., decoders) come from different levels of the feature pyramid that differ in scale and contain different spatial and semantic information and that the learners are independent, our method can be interpreted as a form of self-ensemble segmentation. Usually, the learners of an ensemble must be trained independently. In this work, we show that, in the context of semantic segmentation, this condition can be relaxed and imposed solely to the decoders. Our experiments show that -- all else being equal -- this strategy improves UperNet performance. However, increasing the number of decoders/learners inevitably increases parameters number. Overall, our observations on self-ensemble performance effectiveness but parameter burden, lead us to design a transformer-based model: SenFormer (\textbf{S}elf-\textbf{en}semble segmentation trans\textbf{Former}). Our motivation for using transformer-based learners is that besides transformers' ability to capture long-range dependencies, it has been observed \cite{jaegle2021perceiver,dehghani2019universal,lan2020albert} that recursively applying the same transformer block to the same input features can produce similar -- if not better -- results than using different blocks while reducing the number of parameters and overfitting. Ultimately, our method has fewer parameters and FLOPs than UperNet and performs better.\\

Overall, our SenFormer approach achieves excellent results on various benchmark datasets. Specifically, it outperforms similar architectures \cite{uper2018} that use "feature fusion" strategy, suggesting that our self-ensemble approach effectively leverages the expressive power of ensemble methods. In particular, SenFormer achieves  51.5 mIoU on the benchmark dataset COCO-Stuff-10K \cite{caesar2018cocostuff} and 64.0 mIoU on Pascal-Context \cite{pascal}, outperforming the previous state-of-the-art by a large margin of 6 mIoU and 3.0 mIoU respectively. SenFormer is also on par with state-of-the-art methods on Ade20K \cite{ade20k} and Cityscapes \cite{cordts2016cityscapes}. To summarize, our contributions are two fold:
 \begin{itemize}
     \item We propose an innovative way to leverage the multi-scale features produced by the FPN to form an ensemble of learners inside a single model.
     \item We develop a light-weight transformer-based decoder that is used as a learner in our self-ensemble.
 \end{itemize}

\begin{figure*}[t]
  \centering
  \includegraphics[width=\textwidth]{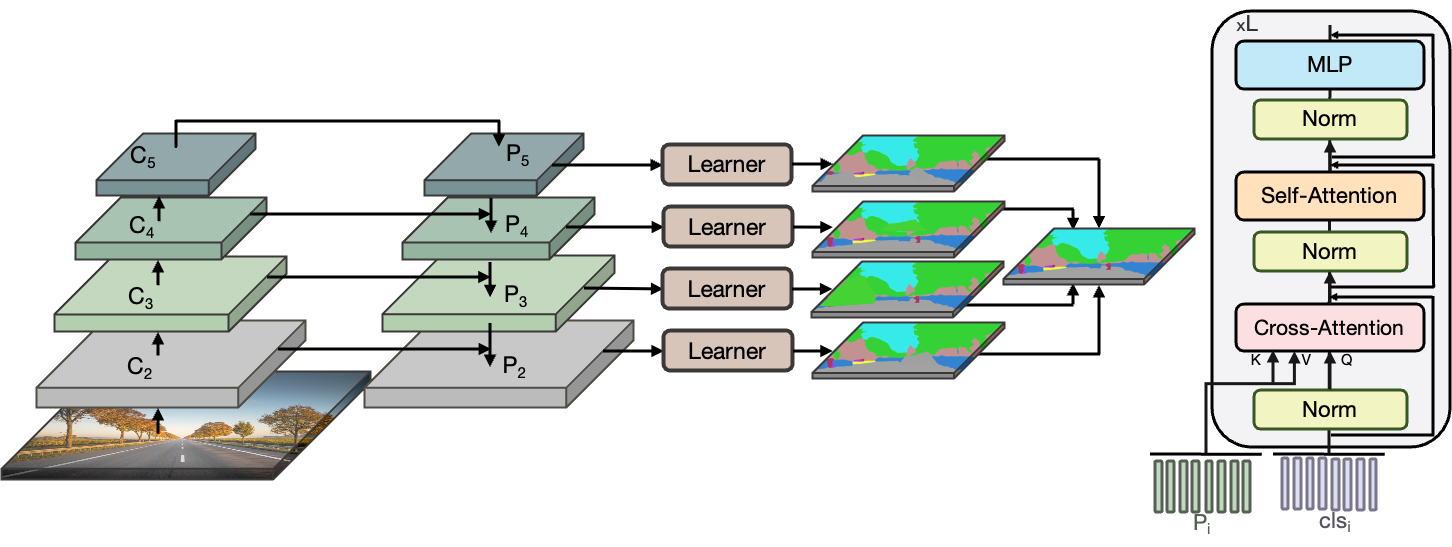}
  \caption{\textbf{SenFormer architecture.} (Left): The features extracted by the backbone $\{C_2,C_3,C_4,C_5\}$ are enhanced in a feature pyramid to produce spatially and semantically strong features maps at every level of the pyramid $\{P_2,P_3,P_4,P_5\}$. Each set of features is decoded by a different learner in the ensemble and the learners' predictions are merged. (Right): architecture of the transformer block.}
  \label{fig:senformer}
\end{figure*}

\section{Related Works}

\textbf{Semantic Segmentation.} Since fully convolutional networks (FCN) have been introduced in the seminal work \cite{FCN14}, it has dominated the field of semantic segmentation. However, due to the inherent locality of the convolution operation, architectures solely based on convolutions struggle to capture long-range dependencies, making it difficult to deal with large and occulted objects. Follow up methods to alleviate the locality issue include but are not limited to atrous convolution \cite{chen2017deeplab,chen2017rethinking}, pyramid  pooling  module \cite{zhao2017pyramid} or the use of FPN \cite{uper2018,kirillov2019panoptic}.
More recently, motivated by the stupendous success of transformer-based architecture for image classification \cite{dosovitskiy2021image,touvron2021training}, multiple works proposed to leverage the self-attention operation to improve segmentation performance of FCN scheme or even completely replace it.
Transformer-based architectures can be used as a drop-in replacement of traditional CNN backbones to enhance the extracted features supplied to the decoder \cite{liu2021swin,touvron2021training,wang2021crossformer,elnouby2021xcit,wang2021pyramid}. It has been observed that transformer backbones that produce a hierarchical feature representation \cite{liu2021swin,wang2021pyramid} are the most suited for segmentation tasks. Additionally, following the original Encoder-Decoder Transformer \cite{vaswani2017attention} used in NLP, recent works proposed to formulate the problem of semantic segmentation as a sequence-to-sequence problem \cite{zheng2021rethinking,strudel2021segmenter,xie2021segformer}, freeing the architecture from any inductive prior biases.

 Alternatively, motivated by the success of DETR \cite{carion2020endtoend} which used transformer for object detection, MaX-DeepLab \cite{wang2021maxdeeplab} and MaskFormer \cite{cheng2021perpixel} treated semantic (and panoptic) segmentation no longer as a per-pixel classification but as a mask classification problem. 
 These architectures first generate a set of candidate masks that are then classified. 
 Shifting the semantic segmentation paradigm from a per-pixel classification to a mask classification problem helps bridge the gap between detection/panoptic segmentation methods and semantic segmentation. It also involves the computation of an assignment score between each generated mask and every class, therefore transferring a part of the training burden to the loss calculation.
 Since our investigation focuses on the efficacy of features fusion and self-ensemble, we will limit our comparisons to per-pixel classification-based architectures.
 
 \textbf{Efficient Ensemble.}
A major limiting factor for building an ensemble of deep learning models is the computational cost during training and testing. Diverse methods were proposed to tackle this issue. By repeatedly applying dropout at inference on an already trained model, Monte Carlo Dropout \cite{gal2016dropout} allows getting many predictions from a single model, ultimately improving its accuracy. BatchEnsemble \cite{wen2019batchensemble} significantly lower ensemble cost by defining each learner's weights to be the Hadamard product between a shared matrix and a rank-one matrix per learner. Snapshot \cite{huang2017snapshot} train a single model to converge to several local minima by leveraging cyclic learning rate scheduling. Other methods for classification include MIMO \cite{havasi2021training}, hyper-batch ensemble \cite{wenzel2020hyperparameter}, late-phase weights \cite{von2020neural} or FGE \cite{garipov2018loss}. For segmentation, \cite{tao2020hierarchical} improves the widely used multi-scale inference by learning relative attention between the scales during training and is used at test-time to greatly improve the performance.
 However, these methods still require several forward passes of the same image, let it be for training or testing. Perhaps most related to our work is TreeNet \cite{lee2015m}, which uses multiple classifier branches that share their early layers. 
 Nevertheless, besides being for classification, unlike to our work, all the learners receive the same input, limiting the depth of the shared part. Moreover, in SenFormer, the parameter cost of the ensemble is further reduced through weight sharing within a learner.


\section{Method}
In this section, we first present the general framework of our method based on self-ensemble as shown in figure \ref{fig:senformer}. Then we detail the different merging strategies. Finally, we describe learners' architecture and the different weight sharing strategies.

Following notations in \cite{he2015deep,lin2017feature,uper2018}, we denote $C_i \in \mathcal{R}^{d_i\times \frac{H}{2^i}\times \frac{W}{2^i}}$ the output of the i-th stage of the bottom-up network (\textit{i.e.} backbone) which has stride of $2^i$ pixels with respect to the input image, where $H\times W$ is the spatial dimension of the input image and $d_i$ the number of channels. Similarly, we denote $P_i \in \mathcal{R}^{d \times \frac{H}{2^i}\times \frac{W}{2^i}}$ the output of the i-th stage of the top-down network (\textit{i.e.} output of the FPN), where $d$ is the numbers of channels in all the feature maps of the FPN. We denote $N$ the number of class.

\subsection{Self-Ensemble}
In this paper, we approach the problem of semantic segmentation as that of a per-pixel classification. Therefore, learners predictions and the merging strategies will be described for an arbitrary pixel and can easily generalize to the whole segmentation map.

An ensemble traditionally consists of $M$ independently trained models called learners. For a given pixel, let denote $X_i\in \mathcal{R}^{N}$ the random variable parameterized by the output of the i-th learner for that particular pixel, which can be decomposed in:
\begin{align}
  X_i & = Y + \epsilon_i
\end{align}
where $Y$ is the target and $\epsilon_i$ is the prediction error of the i-th learner.

The most straightforward way to merge  different learners' predictions is by averaging them. It is well known \cite{opitz1999popular} \cite{zhou2002ensembling} \cite{kotsiantis2006machine} that the ensemble performance is usually better than the individual learners.

Classical statistics suggest that when the predictions are roughly independent, the last term in equation \ref{eq:var} is close to zero and therefore averaging greatly reduces the noise.

\begin{align} \label{eq:var}
    Var(\frac{1}{M} \sum_{i=1}^M \epsilon_i) =\frac{1}{M^2} \sum_{i=1}^M Var(\epsilon_i) + \frac{2}{M^2} \sum_{i<j} Cov(\epsilon_i, \epsilon_j).
\end{align}

On another note,  a recent study suggests that this hypothesis might not hold in the context of deep learning. In \cite{allenzhu2021understanding}, Allenzhu \textit{et. al,} acknowledge that for the task of image classification, the different learners learn to detect different views/features of the object of interest depending on their weight initialization. However, there are some images taken from a particular angle where the learned features may be missing. Hence, when the ensemble is large enough, all possible views are captured, thus increasing the model's accuracy. Note, however, that it is not clear in \cite{allenzhu2021understanding} if this result also holds for semantic segmentation.
Either way, a key requirement is that the learners' predictions must be independent, let it be for the variance reduction or the multi-view hypothesis.

Total independence of the predictions implies tediously training multiple independent models. 
In this paper, we aim at relaxing the independence hypothesis to reduce the training cost, while maintaining the performance benefits of ensemble. To do so, the learners/decoders share the same backbone but receive input features coming from different levels of the feature pyramid, \textit{i.e.}, $\{ P_2, P_3, P_4, P_5 \}$, as shown in figure \ref{fig:senformer}. 

Nevertheless, it is observed that if one trains the different learners of an ensemble altogether (i.e., applying the loss on the merged prediction), the performance boost offered by the ensemble disappears \cite{allenzhu2021understanding}. However, we show in our experiments that it is not the case in our setting. We hypothesize that it is because each learner is independently initialized (as in ensemble) and receives different inputs, therefore alleviating the need for separate training. In this manner, several segmentation predictions can be obtained with only a single forward pass of the input image.

\subsection{Merging strategies}
We describe the different methods considered to merge the different learner predictions (during inference). 

\textbf{Averaging.} It is the most commonly used method for prediction merging as no additional trainable parameters are required. The merged prediction $X_{avg}$ of $M$ learners is obtained by:
\begin{align}
    X_{avg} = \frac{1}{M} \sum_i X_i
\end{align}

\textbf{Product.} The predicted probability for each pixel is multiplied rather than average. That way, more weight is given to learners with high confidence.
The merged prediction $X_{prod}$ of $M$ is given by:
\begin{align}
    X_{prod} = \prod_{i=1}^M  X_i
\end{align}

\textbf{Majority vote.} Each learner assigns a vote to the class with the largest confidence. The merged prediction $X_{maj}$ is obtained by:
\begin{align}
    X_{maj} = \frac{1}{M} \sum_i \delta_c(X_i) \quad \text{where} \quad c = \underset{1 \leq j \leq N}{argmax} \  X_i^{j}
\end{align}
where $\forall j \leq N, \delta_c(X_i)^j = \left\{
    \begin{array}{ll}
        X_i^j & \mbox{if } j=c \\
        0 & \mbox{else.}
    \end{array}
\right.$

\textbf{Hierarchical Attention.} We borrow the "attention module" from \cite{tao2020hierarchical} that is used to learn a relative attention mask between adjacent scales. The module consists of $(3\times 3 \, conv) \rightarrow (BatchNorm)
\rightarrow (ReLU) \rightarrow (3\times 3 \, conv) \rightarrow (BatchNorm) \rightarrow (ReLU) \rightarrow (1\times 1 \, conv) \rightarrow (Sigmoid)$, where the last convolution output a single (attention) map. In the original paper, the module is fed with the same input features maps of the decoder. Another variant would be to use the segmentation logits (decoder's output) instead. In our experiment, we tried both and found the latter to work better with SenFormer. Since SenFormer has four learners, we need 3 "attention modules" to predict the relative attention maps.

\textbf{Explicit Attention.} We used the same "attention module" as for Hierarchical Attention \cite{tao2020hierarchical}, but trained it to predict a dense mask for each scale rather than a relative mask.\\

Surprisingly, our experiments found the simple "averaging" strategy to perform better than others, except for the "hierarchical attention" (Table \ref{tab:merging}). However, given the performance boost of the "attention module" is limited, it does not justify the overhead complexity.
Therefore, SenFormer uses the "averaging" as the default merging strategy since it yields high performance without requiring additional parameters.

\subsection{Learner architectures}
Hereafter, we described the architecture of a single learner/decoder.

As depicted in Figure \ref{fig:senformer}, the $i^{th}$ decoder branch takes as input the features coming from the corresponding level of the FPN (with stride $s_i$) $P_i \in \mathcal{R}^{d \times \frac{H}{2^i} \times \frac{W}{2^i}}$, as well as a set of $N$ learnable embeddings termed as class embeddings, $ \textbf{cls}_i = [cls^1_i, \dots, cls^{N}_i] \in \mathcal{R}^{N \times d}$, where $N$ is the number of class. In this respect, there is one learnable class embedding $cls^k_i$ per segmentation class and per level in the feature pyramid.

Each decoder is a transformer composed of $L$ layers whose architecture is inspired by the traditional transformer \cite{vaswani2017attention}. Note however that a "pre-norm" strategy is used in place of "post-norm" for the placement of Layer Normalization (LN), \textit{i.e.}, the skip connections inside each transformer block are not affected by the LN \cite{nguyen2019transformers} (see ablation study in the Annex). 

In a nutshell, a single Transformer Decoder block consists of three successive operations: Cross-Attention, Self-Attention and Multi-Layer Perceptron layers. In the Cross-Attention operation the feature map $P_i$ is used as key and value while the class embedding $\textbf{cls}_i$ is used a query. The Self-Attention and MLP are applied only to the class embeddings.

Finally, each decoder/learner is composed of L layers of decoder block and its prediction is obtained via a dot product between the class embeddings $\textbf{cls}_i$ and the corresponding feature pyramid feature $P_i$ -- see the Annex for more details. However, using multiple decoders greatly increases the number of parameters. To mitigate this, we explore several weight-sharing strategies.

\subsection{Weight sharing}
Weight sharing is a commonly used technique to reduce the number of parameters \cite{jaegle2021perceiver,dehghani2019universal,lan2020albert}, while also regularizing the optimization by reducing the degree of freedom which mitigates overfitting. However, in the context of Ensemble, special care 
regarding the kind of weight sharing used must be given.

Two types of weight sharing can be used: inter-learner and intra-learner sharing.
The former involves sharing parameters between the different learners, while latter within the learner.
Figure \ref{fig:WeightSharing} depicting the different sharing methods can be found in Annex. 

\textbf{Repeated block.} A given learner is composed of a single decoder block recursively used $L$ times. It is a form of "intra-learner sharing" since no parameters are shared between the different learners.

\textbf{Decoder sharing.} The different learners share the same decoder but have their own class embedding. It is a form of "inter-learner sharing".

\textbf{Class embeddings sharing.} The same learnable class embeddings \textbf{cls} is used for all the learners. It is also a form of "inter-learner sharing".\\

Table \ref{tab:WeightSharing} shows that any "inter-learner sharing" strategy significantly degrades the segmentation performance, confirming the importance of keeping the different learners as independent as possible. Conversely, the "repeated block" strategy performs better than when no sharing is used, while significantly reducing the number of parameters. Hence, SenFormer uses the "repeated block" as the default weight sharing policy.

\section{Experiments}

\textbf{Datasets.}
We evaluate our model performance using four semantic segmentation benchmark datasets, ADE20K \cite{ade20k}, Pascal Context \cite{pascal}, COCO-Stuff-10K \cite{caesar2018cocostuff} and Cityscapes \cite{cordts2016cityscapes}.
We use ADE20K, which is a challenging scene parsing dataset consisting of 20,210 training images and 2,000 validation images and covers $150$ fine-grained labeled classes, for the ablation studies. Please see the Annex for detailed descriptions of all used datasets.

\textbf{Evaluation metric.}
We report the mean Intersection over Union (mIoU), a standard metric for semantic segmentation.

\textbf{Baseline model.}
To demonstrate that the performance improvement of our method is genuinely a result of self-ensemble instead of feature fusion, we introduce a simple decoder baseline module that borrows the features fusion strategy from UperNet \cite{uper2018}, but uses our transformer decoder-- see Figure \ref{fig:featurefusion}. This way, the FeatureFusionBaseline and SenFormer only differ by the multi-scale fusion strategy. Following \cite{uper2018}, the baseline multi-scale fusion strategy is as follow: we first resize (through bilinear interpolation) all the features $\{P_2,P_3,P_4,P_5\}$ to match $P_2$ dimension (\textit{i.e.} $1/4$ of the input image) and concatenate them. We then apply a $3\times 3$ convolution followed by a batch normalization layer and a ReLU activation.
Note that this baseline is only used for ablation purposes and SenFormer is thereafter also compared to state of the art methods in Section \ref{sec:resSOTA}.

\begin{table}{}


\caption{Performances by using different learner combinations, where \ding{51}/\ding{55} indicates whether the learner is used for the prediction.}\label{tab:ens?}
\centering
\begin{tabular}{ccccc}
\toprule

 $d_2$ \quad & $d_3$ \quad & $d_4$ \quad & $d_5$ \quad & \textit{mIoU} \quad\\
\midrule[\heavyrulewidth]

\ding{51} & \textcolor{gray}{\ding{55}} & \textcolor{gray}{\ding{55}} & \textcolor{gray}{\ding{55}}& 42.90\\
\textcolor{gray}{\ding{55}} & \ding{51} & \textcolor{gray}{\ding{55}} & \textcolor{gray}{\ding{55}}& 42.08\\
\textcolor{gray}{\ding{55}} & \textcolor{gray}{\ding{55}} & \ding{51} & \textcolor{gray}{\ding{55}}& 41.40\\
\textcolor{gray}{\ding{55}} & \textcolor{gray}{\ding{55}} & \textcolor{gray}{\ding{55}} & \ding{51}& 38.23\\
 \ding{51} & \ding{51} & \ding{51}& \textcolor{gray}{\ding{55}} & 44.12\\
\rowcolor{almond} \ding{51} & \ding{51} & \ding{51} & \ding{51}& \textbf{44.38}\\
\bottomrule
\end{tabular}
\end{table}
\begin{table}
\caption{Performance comparisons on ADE20K validation of different weight sharing settings for SenFormer.$^\clubsuit$ indicates SenFormer's default setting.}\label{tab:WeightSharing} 
\centering
\begin{tabular}{c c c c c }
\toprule
weight sharing & \multirow{2}{*}{\# blocks} & \multirow{2}{*}{\textit{mIoU}} & \multirow{2}{*}{\#params.}& \multirow{2}{*}{FLOPs}    \\
setting & & & &\\
\midrule[\heavyrulewidth]
decoder shared & 6 & 42.69 & 68M  & 179G  \\
cls embeddings & 6 & 42.91 & 67M & 179G  \\
\rowcolor{almond} repeated$^\clubsuit$ & 6 & 44.38 & \textbf{55M} & 179G  \\
none & 1 & 43.12 & \textbf{55M} & \textbf{111G}  \\
 none & 6 & \textbf{44.68} & 144M & 179G\\
\midrule
\midrule
\multicolumn{2}{c}{UperNet} & 42.05 & 67M & 238G  \\
\bottomrule
\end{tabular}
\end{table}

\subsection{Implementation and training details}
\textbf{Backbones.} Since SenFormer uses the FPN to build a multi-scale set of features, it is compatible with any backbone architecture. In our experiments we use both convolutional ResNet50 and ResNet101 \cite{he2015deep} and the different size of the transformer-based backbones Swin-Transformers\cite{liu2021swin}.

\textbf{FPN.}  The channel dimension of the feature pyramid $d$ is set to $512$. For small backbones, we find it beneficial in the FPN to replace the traditional $3\times 3$ convolution by a Window-based Transformer Block from \cite{liu2021swin}. Since it introduces a marginal computation overhead, we applied it to all backbones. Please see the appendix for detailed ablation of the FPN.

\textbf{Decoders.} 
Each learner is independently supervised with a cross-entropy loss. In addition, we apply a standard cross-entropy loss on the final ensemble prediction. 

\textbf{Training setting.} We use mmsegmentation \cite{mmseg2020} library as codebase and follow the standard training practice for each dataset. Moreover, we apply common data augmentation for semantic segmentation, which include left-right flipping, standard random color jittering, random resize with ratio $0.5-2$ and random cropping.

For the optimizer, we use AdamW \cite{loshchilov2019decoupled}. As common practice for segmentation, we use "\textit{poly}" learning rate scheduler. Following \cite{uper2018,cheng2021perpixel}, we set the initial learning rate to $10^{-4}$ and weight decay to $10^{-4}$ for ResNet backbones, an initial learning rate of $6\cdot10^{-6}$ and a weight decay of $10^{-2}$ for Swin-Transformer. We also use gradient clipping of $3$ to help stabilizing the training, and for ResNet backbones a learning rate multiplier of $0.1$ is applied. 

During training, the input images are cropped to a size of $512\times 512$ for ADE20K \cite{ade20k} and COCO-Stuff-10K \cite{caesar2018cocostuff}, $512\times 512$ for Pascal Context \cite{pascal} and $512\times 1024$ for Cityscapes \cite{cordts2016cityscapes}, unless stated otherwise.
All the models are trained on 8 V100 GPUs with a batch size of 8 for Cityscapes and 16 for the others (see Annex for details). The segmentation performance is reported using single-scale inference. 
Finally, all the backbones are pretrained on ImageNet-1K \cite{russakovsky2015imagenet} unless stated otherwise.

\subsection{Self-ensemble vs Features Fusion}
Features at different levels of the pyramid carry different scale of contextual information, and our experiments support that self-ensemble is capable of capturing and integrating such information.


\begin{table}
\caption{Experiments with different number of decoder blocks.}\label{tab:num_block}
\centering
\begin{tabular}{cccc}
\toprule
\# block & \textit{mIoU} & \#params. & FLOPs \\ 

 \midrule[\heavyrulewidth]

1 & 42.44 & 55M & 111G \\
3 & 43.25 & 55M & 139M \\
\rowcolor{almond} 6 & 43.6 & 55M & 179G \\
9 & \textbf{43.7} & 55M & 220G \\
\bottomrule
\end{tabular}
\end{table}

\begin{table}{}
\caption{Comparisons of the features fusion and self-ensemble strategies.$^\clubsuit$indicates self-ensemble.}
\label{tab:FFvsSEn}
\centering
\begin{tabular}{c c c c }
\toprule
method & \textit{mIoU} & \#params.& FLOPs    \\
\midrule[\heavyrulewidth]
UperNet & 42.02 & 67M &238G  \\
SenUperNet$^\clubsuit$ & 42.8 & 70M & \textbf{135G} \\
FeaturesFusionBaseline & 43.1 & \textbf{52M} &307G  \\
\rowcolor{almond} SenFormer$^\clubsuit$ & \textbf{44.3} & 55M & 179G\\
\bottomrule
\end{tabular}
\end{table}


\textbf{Ensemble effect.}
 We first analyze the output produced by each decoder and assess their performance. Table \ref{tab:ens?} outlines the \textit{mIoU} scores of independent prediction of each decoder as well as for the ensemble. Notably, the ensemble \textit{mIoU} score is $+3.5$ better than the mean score of the learners taken separately with $\frac{1}{4}\sum_{i=2}^{5} \textit{mIoU}(d_i) = 41.15$. More surprisingly, even though $d_5$ taken separately performs significantly worse than the others -- due to its low-resolution inputs -- it positively contributes to the ensemble, consistent with traditional ensemble methods where even weak learners can be combined to enhance the overall prediction.

\textbf{Does the performance boost really comes from self-ensemble?} 
To rule out the performance gain brought by the use of transformer-based decoders rather than convolution, we compare SenFormer and the FeaturesFusionBaseline, since they only differ in the multi-scale fusion strategy (\emph{features fusion vs. self-ensemble}). In Table \ref{tab:FFvsSEn}, we observe that SenFormer is +$2$ \textit{mIoU} better than the baseline. Conversely, we applied the self-ensemble method to UperNet \cite{uper2018} by using the same convolution-based decoder at each level of the feature pyramid rather than merging the features. Likewise, the self-ensemble version (SenUperNet) performs better than the vanilla UperNet, suggesting that our self-ensemble approach is the main driver for improvement.

\begin{table*}
  
  \caption{Self-ensemble SenFormer vs features fusion UperNet on ADE20K validation. Backbones pre-trained on ImageNet-22K are marked with $^\ddagger$.}\label{tab:main_res}
  \centering
  \begin{tabular}{l|llllll}
    \toprule
     & method & backbone  &crop size& \#params. &FLOPs & \textit{mIoU}\\
    \midrule
    
\multirow{2}{*}{\rotatebox[origin=c]{90}{CNN}}  &  UperNet     & ResNet-50 & 512$\times$512 &67M& 238G& 42.05\\
     & \cellcolor{almond}  SenFormer     &\cellcolor{almond} ResNet-50 & \cellcolor{almond} 512$\times$512 & \cellcolor{almond} \textbf{55M} & \cellcolor{almond}\textbf{179G} & \cellcolor{almond} \textbf{44.38} \\
     &  UperNet     & ResNet-101 & 512$\times$512 &86M&  257G&  43.82\\
          & \cellcolor{almond}   SenFormer     & \cellcolor{almond} ResNet-101 & \cellcolor{almond} 512$\times$512 & \cellcolor{almond} \textbf{79M} & \cellcolor{almond} \textbf{199G}& \cellcolor{almond} \textbf{46.93}\\
    \midrule
    \midrule
     \multirow{4}{*}{\rotatebox[origin=c]{90}{Transformer}} & UperNet     & Swin-T & 512$\times$512 &60M& 236G& 44.41 \\
     & \cellcolor{almond}  SenFormer     & \cellcolor{almond} Swin-T & \cellcolor{almond} 512$\times$512 & \cellcolor{almond} \textbf{59M} & \cellcolor{almond} \textbf{179G}& \cellcolor{almond}  \textbf{46.0}\\
     & UperNet     & Swin-S & 512$\times$512 &81M& 259G & 47.72\\
     & \cellcolor{almond}  SenFormer     & \cellcolor{almond} Swin-S & \cellcolor{almond} 512$\times$512 & \cellcolor{almond} 81M& \cellcolor{almond} \textbf{202G} & \cellcolor{almond} \textbf{49.2}\\
     & UperNet     & Swin-B$^\ddagger$ & 640$\times$640 &121M& 471G& 50.04\\
     & \cellcolor{almond}  SenFormer    & \cellcolor{almond} Swin-B$^\ddagger$ & \cellcolor{almond} 640$\times$640 & \cellcolor{almond}\textbf{120M}& \cellcolor{almond} \textbf{371G}& \cellcolor{almond} \textbf{52.21}\\
     & UperNet     & Swin-L$^\ddagger$ & 640$\times$640 & 234M& 647G & 52.05\\
     & \cellcolor{almond} SenFormer    & \cellcolor{almond} Swin-L$^\ddagger$ & \cellcolor{almond} 640$\times$640 & \cellcolor{almond}\textbf{233M}& \cellcolor{almond} \textbf{546G} & \cellcolor{almond}  \textbf{53.08}\\
    \bottomrule
  \end{tabular}
\end{table*}

\textbf{SenFormer vs UperNet.}
We compare SenFormer with UperNet architecture for a variety of CNN- and transformer-based backbones. As we can see from Table \ref{tab:main_res}, when using the same standard Swin-Transformer backbone, SenFormer consistently outperforms UperNet regardless of the backbone size. The performance gap is even larger when using convolutional backbones (+$3$ \textit{mIoU}), suggesting that our transformer-based decoder successfully captures the long-range dependencies missed by the CNN-based backbones.

Thanks to its weight sharing strategy, SenFormer has fewer parameters than UperNet. Furthermore, since SenFormer avoids the computationally expensive features merging operation, it also has substantially fewer FLOPs.

\subsection{Comparison to state-of-the-art}\label{sec:resSOTA}
In this section we further compare SenFormer to state-of-the-art methods on ADE20K and additional benchmark datasets.

\begin{table}
\caption{Benchmark on ADE20K validation set.}\label{tab:add-ade}
\centering
\begin{tabular}{c c c c }
\toprule
method & backbone & \textit{mIoU} & +MS    \\
\midrule[\heavyrulewidth]
DeepLabV3+ \cite{chen2018encoder} & R50 & 44.0 & 44.9  \\
PerPixelBaseline+\cite{cheng2021perpixel} & R50 & 41.9 & 42.9\\
MaskFormer \cite{cheng2021perpixel} & R50 & \textbf{44.5} & \textbf{46.7}\\
\rowcolor{almond} SenFormer & R50 & 44.4 & 45.2\\
\midrule
OCRNet \cite{yuan2021segmentation} & R101 & - & 45.3\\
DeepLabV3+ \cite{chen2018encoder}& R101 & 45.5 & 46.4\\
MaskFormer \cite{cheng2021perpixel}& R101 & 45.5 & 47.2\\
\rowcolor{almond} SenFormer & R101 & \textbf{46.9} & \textbf{47.9}\\
\midrule
\midrule
SETR-L  MLA \cite{zheng2021rethinking}& ViT-L & - & 50.3 \\
Segmenter \cite{strudel2021segmenter}& ViT-L & 50.71 & 52.25  \\
Segmenter-Mask\cite{strudel2021segmenter}& ViT-L & 51.82 & 53.63 \\
SegFormer \cite{xie2021segformer}& MiT-B5 & 51.0 & 51.8  \\
UperNet \cite{liu2021swin}& Swin-L & 52.05 & 53.5\\
\rowcolor{almond} SenFormer & Swin-L & 53.08 & 54.2\\
MaskFormer \cite{cheng2021perpixel} & Swin-L & \textbf{54.1} & \textbf{55.6}\\ 
\bottomrule
\end{tabular}
\end{table}

\begin{table}{}
\caption{Benchmark on Pascal Context test. $^\ddagger$/\textcolor{blue}{blue} indicates previous/new SOTA.}\label{tab:add-pascal}
\centering
\begin{tabular}{c c c c }
\toprule
method & backbone & \textit{mIoU} & +MS    \\
\midrule[\heavyrulewidth]
DANet \cite{fu2019dual} & R50 & - & 50.5  \\
EMANet \cite{li2019expectation} & R50 & - & 50.5  \\
CAA \cite{huang2021channelized}& R50 & 50.23 & -\\
\rowcolor{almond} SenFormer & R50 & \textbf{53.18} & \textbf{54.3}\\
\midrule
DANet \cite{fu2019dual} & R101 & - & 52.6  \\
EMANet \cite{li2019expectation} & R101 & - & 53.1  \\
DeepLabV3+ \cite{chen2018encoder}& R101 & 53.2 & 54.67\\ 
OCRNet \cite{yuan2021segmentation}& R101 & - & 54.8\\
CAA \cite{huang2021channelized}& R101 & - & 55.0\\
\rowcolor{almond} SenFormer & R101 & \textbf{54.6} & \textbf{56.6}\\
\midrule
OCRNet \cite{yuan2021segmentation}& HRNet & - & 56.2\\
CAA \cite{huang2021channelized}& EN-B7 & 58.40 & 60.5$^\ddagger$\\
\midrule
\midrule
SETR-L MLA\cite{zheng2021rethinking}& ViT-L & 54.9 & 55.8 \\
Segmenter \cite{strudel2021segmenter}& ViT-L & 58.1 & 59.0 \\
\rowcolor{almond} SenFormer & Swin-L & \textbf{63.1} & \textcolor{blue}{\textbf{64.5}}\\
\bottomrule
\end{tabular}
\end{table}

\textbf{ADE20K.} In Table \ref{tab:add-ade} we compare SenFormer to a variety of FCN- and transformer-based decoders using both CNN- and transformer-based backbones. When using standard ResNet backbones, SenFormer outperforms all other methods. The same can be said for per-pixel classification-based models when using transformer-based backbones, where SenFormer even outperforms recently introduced transformer-based decoders like SETR \cite{zheng2021rethinking}, Segmenter \cite{strudel2021segmenter} and SegFormer \cite{xie2021segformer}. Note however that MaskFormer \cite{cheng2021perpixel} is doing better than SenFormer when using transformer-based backbones. Indeed, MaskFormer introduces a new approach for semantic segmentation that is based on mask classification (rather than traditional per-pixel classification) and that greatly improves segmentation performances. In fact, MaskFormer \cite{cheng2021perpixel} significantly outperforms PerPixelBaseline+ \cite{cheng2021perpixel} while sharing the same architecture and only differing by the problem formulation \textit{(per-pixel vs mask classification)}. We plan to formulate SenFormer as mask classification in our future work, as it has significant potential to improve segmentation.

\textbf{Pascal Context.} In Table \ref{tab:add-pascal} we compare SenFormer to current state-of-the-art methods on Pascal Context test dataset, which is obtained by CAA \cite{huang2021channelized} using EfficientNet-B7(EN-B7) as backbone, with a mIoU of $60.5$. SenFormer outperforms previous FCN methods when using standard ResNet backbones, as well as recent transformer-based methods. SenFormer outperforms the current state-of-the-art (CAA) when using the same ResNet-101 backbone, showing the benefit of our approach. Moreover, we reach a score of \textbf{64.0} mIoU when using Swin-L as backbone.  Overall, our approach shows a significant improvement of +3.5 mIoU over the previous state-of-the-art.


\begin{table}
\caption{Benchmark on COCO-Stuff-10K test. $^\ddagger$/\textcolor{blue}{blue} indicates previous/new SOTA.}\label{tab:add-coco}
\centering

\begin{tabular}{c c c c }
\toprule
method & backbone & \textit{mIoU} & +MS    \\
\midrule[\heavyrulewidth]
EMANet \cite{li2019expectation} & R50 & - & 37.6  \\
PerPixelBaseline+\cite{cheng2021perpixel} & R50 & 34.2 & 35.8\\
MaskFormer \cite{cheng2021perpixel} & R50 & 37.1 & 38.9\\
\rowcolor{almond} SenFormer & R50 & \textbf{40.0} & \textbf{41.3}\\
\midrule
DANet \cite{fu2019dual} & R101 & - & 39.7  \\
EMANet \cite{li2019expectation} & R101 & - & 39.9  \\
OCRNet \cite{yuan2021segmentation}& R101 & - & 39.5\\
CAA \cite{huang2021channelized}& R101 & - & 41.2\\
MaskFormer \cite{cheng2021perpixel} & R101 & 38.1 & 39.8\\
\rowcolor{almond} SenFormer & R101 & \textbf{41.0} & \textbf{42.1}\\
\midrule
OCRNet \cite{yuan2021segmentation}& HRNet & - & 40.5\\
CAA \cite{huang2021channelized}& EN-B7 & - & 45.4$^\ddagger$\\
\rowcolor{almond} SenFormer & Swin-L & \textbf{49.8} & \textcolor{blue}{\textbf{51.5}}\\
\bottomrule
\end{tabular}
\end{table}


\textbf{COCO-Stuff-10K.} Table \ref{tab:add-coco} compares SenFormer to state-of-the-art methods on COCO-Stuff-10K test dataset, which is obtained by CAA \cite{huang2021channelized} using EfficientNet-B7(EN-B7) as backbone,  with a mIoU of $45.4$.
When using standard ResNet backbones, SenFormer outperforms previous FCN methods, as well as the transformer-based method MaskFormer. Moreover, we obtain \textbf{51.5} mIoU when using Swin-L as backbone, establishing a new SOTA by a substantial margin of +6 mIoU over previous methods on COCO-Stuff-10K.

\textbf{Cityscapes.} Table \ref{tab:add-cityscapes} compare SenFormer to state-of-the-art methods on Cityscapes validation dataset. We observe that SenFormer performs on par with the best FCN and transformer-based methods. We hypothesis that since Cityscapes dataset has only 19 classes, the object classification aspect of the segmentation is easier and therefore SenFormer cannot benefit as much from its class embeddings, as it does with datasets where the number of classes is larger.

\begin{table}
\caption{Benchmark on Cityscapes val.}\label{tab:add-cityscapes}
\centering
\begin{tabular}{c c c c }
\toprule
method & backbone & \textit{mIoU} & +MS    \\
\midrule[\heavyrulewidth]
MaskFormer \cite{cheng2021perpixel} & R50 & 78.5 & -\\
\rowcolor{almond} SenFormer & R50 & 78.8 & 80.1\\
DeepLabV3+\cite{chen2018encoder}& R50 & \textbf{78.97} & \textbf{80.46} \\ 
\midrule
MaskFormer \cite{cheng2021perpixel} & R101 & 79.7 & 81.4\\
\rowcolor{almond} SenFormer & R101 & 79.9 & 81.4 \\
OCRNet \cite{yuan2021segmentation}& R101 & - & 82.0\\
DeepLabV3+ \cite{chen2018encoder}& R101 & \textbf{80.9} & \textbf{82.03}\\ 
\midrule
SETR-L PUP  \cite{zheng2021rethinking}& ViT-L & - & 82.2 \\
Segmenter \cite{strudel2021segmenter}& ViT-L & - & 80.7  \\
Segmenter-Mask\cite{strudel2021segmenter}& ViT-L & 79.1 & 81.3 \\
\rowcolor{almond} SenFormer & Swin-L & \textbf{82.8} & \textbf{84.0} \\
SegFormer \cite{xie2021segformer}& MiT-B5 & 82.4 & \textbf{84.0}  \\
\bottomrule
\end{tabular}
\end{table}
\subsection{Ablation studies}\label{sec:ablations}
For ablation studies, we solely use ResNet-50 pre-trained on ImageNet-1K \cite{russakovsky2015imagenet} as backbone and trained all models for 100k iterations.

\textbf{Number of blocks.} Table \ref{tab:num_block} show the results of SenFormer trained with varying number of decoder block per learner. The \textit{mIoU} improves with more decoder blocks added. Note that using a single decoder block leads to significantly poorer performance, suggesting that all the information in $P_i$ can not be transfered to the class embeddings $\textbf{cls}_i$ in one operation. 
We choose to use 6 decoder blocks per learner as it offers a good complexity/performance trade-off. 

\textbf{Weight sharing.} Table \ref{tab:WeightSharing} compares different weight sharing approaches for SenFormer. First, we observe that policies that involve sharing weights between the learners ("decoder sharing" and "cls embeddings") lead to a significant drop in performance, even when the number of parameters is not reduced. For example, "cls embedding" is 1.2mIoU lower than the base setting while having 89M more parameters, confirming that independence between learners is a key component in SenFormer. Furthermore, recursively applying the same decoder block leads to non trivial performance boost of +1mIoU, while having the same parameter number.

\begin{table}{}

\parbox{.4\linewidth}{
\caption{Merging strategies.}\label{tab:merging}

\centering
    \begin{tabular}{c  c}
    \toprule
         merging strategy    & mIoU \\
    \midrule[\heavyrulewidth]
\rowcolor{almond} averaging & 44.4 \\
        product & 40.28 \\
        majority vote & 39.89 \\
        hierarchical att. & 44.5 \\
        explicit att. & 39.7 \\
    \bottomrule
    \end{tabular}}
\hfill 
\parbox{.4\linewidth}{
\caption{Ensemble and learners variance on Ade20K validation.}\label{tab:variance}
\centering
    \begin{tabular}{c  c}
    \toprule
        Output     & var. $(10^{-3})$ \\
    \midrule[\heavyrulewidth]
        ensemble & 56.6 \\
        $d_2$ & 56.0 \\
        $d_3$ & 56.8 \\
        $d_4$ & 56.3 \\
        $d_5$ & 55.2 \\
    \bottomrule
    \end{tabular}}
    \end{table}
\begin{table}
\caption{Effect of increasing the number of learners.}\label{tab:multiLearner}
\centering
\begin{tabular}{c  c c}
    \toprule
    \# learners & total & \multirow{2}{*}{\textit{mIoU}}\\
per scale & \# learner & \\
    \midrule[\heavyrulewidth]
   \rowcolor{almond}     1 & 4 & \textbf{44.3} \\
        2 & 8 & 44.2 \\
    \bottomrule
    \end{tabular}
    
\end{table}


\section{Discussion}
\textbf{Variance reduction.}
A common explanation for the better performance of the ensemble over its composing elements is that by averaging the variance over the merged prediction is reduced. To test this assumption, for each input image in the Ade20K validation set, we computed for the ensemble and for each learner the variance over the segmentation map prediction for each pixel (\textit{i.e.}, the variance along the channel axis). We then averaged over the entire validation set.
As shown in Table \ref{tab:variance}, the ensemble variance is not significantly smaller than the variance of the individual learners. Consequently, the variance reduction interpretation may not apply in the context of self-ensemble, and more broadly for deep learning models \cite{allenzhu2021understanding}.\\

\textbf{Multi-view approach.}
A more recent explanation for the success of Ensemble is that the different learners capture multi-views present in the data \cite{allenzhu2021understanding}. However, since the mutli-scale inputs of the learners come from the same backbone, it is very unlikely that they focus on different views of the objects of interest. We rather hypothesize that in SenFormer the boost in performance does not emerge from the different random initialization of the learners that will learn to focus on specific views of the input image, but rather from the different scale information captured by the FPN. Consequently, using more than one learner per level in the feature pyramid will not yield better results. It is indeed confirmed by results in Table \ref{tab:multiLearner} where SenFormer performances do not improve with additional learners.


\section{Conclusions}
This paper introduces our self-ensemble approach for semantic segmentation, a simple methodology that benefits from ensemble learning while avoiding the inconvenience and cost of training multiple times the same model. We leveraged the multi-scale feature set produced by FPN-like methods to build an ensemble of decoders within a \emph{single model}, where learners in the ensemble are fed with features coming from different levels of the feature pyramid. We also developed a transformer-based architecture for the learner/decoders.

Our approach outperforms current state-of-the-art on Pascal Context and COCO-Stuff-10K datasets and is competitive on Ade20K and Cityscapes datasets for semantic segmentation. It is more efficient in terms of FLOPs and limit the number of parameter thanks to weight sharing. It is part of our future work to investigate "mask classification" formulation of semantic segmentation.

\section*{Acknowledgements}
We thank Romain Fabre for insightful discussion without which this paper
would not be possible. This work was partially supported by the National Institutes of Health (NIH), National Cancer Institute (NCI) Human Tumor Atlas Network (HTAN) Research Center (U2C CA233280), and and a NIH/NCI Cancer Systems Biology Consortium Center (U54 CA209988).

{\small
\bibliographystyle{ieee_fullname}
\bibliography{egbib}
}

\appendix

\begin{figure*}[t!]
    \centering
    
    \includegraphics[width=0.9\linewidth]{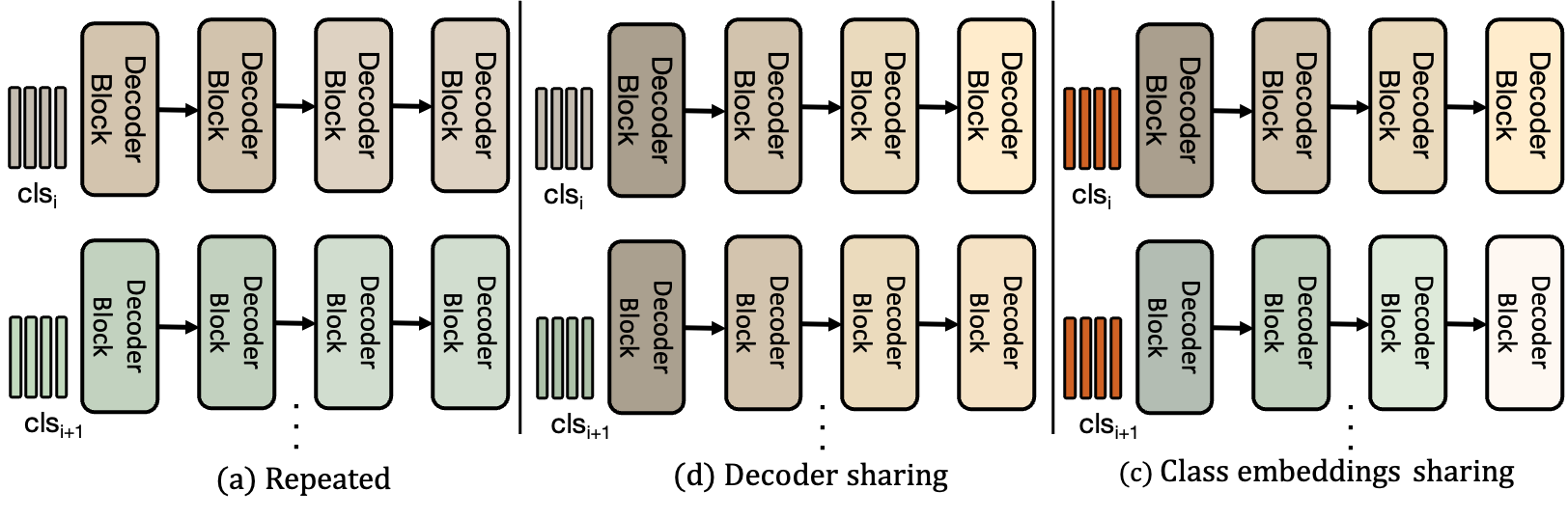}
    \caption{Weight sharing policies. Different color indicates different set of parameters.} \label{fig:WeightSharing}
\end{figure*}
\vspace*{4in}

\section*{Supplementary Materials}

We first provide more information about the datasets used to evaluate SenFormer performances (Section \ref{sec:add-datasets}).
In an attempt to gain more insights about our self-ensemble approach, additional experiments and discussions on SenFormer are presented. Eventually, we provide qualitative segmentation results of SenFormer.

  \begin{figure*}[] 
    \centering
    
    \includegraphics[width=1\linewidth]{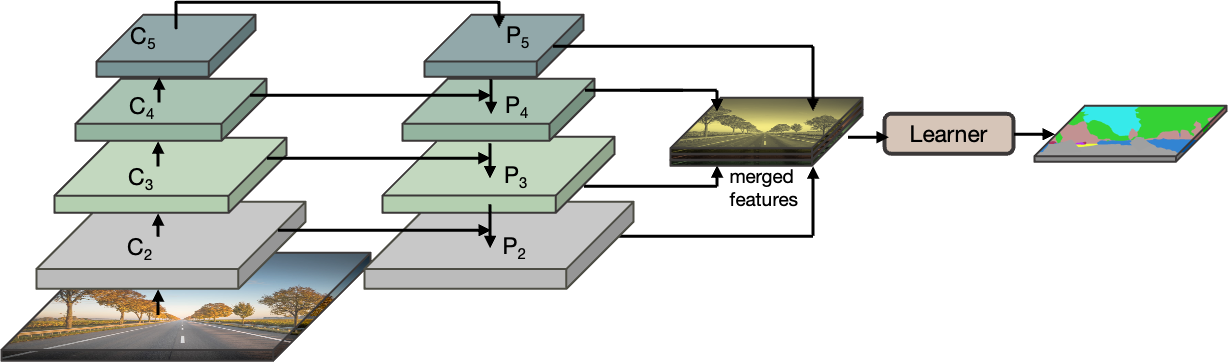}
    \caption{FeaturesFusionBaseline architecture.} \label{fig:featurefusion}
\end{figure*}

\section{Datasets Descriptions}\label{sec:add-datasets}

\textbf{ADE20K} \cite{ade20k} is a scene parsing dataset built from ADE20K-Full dataset, where $150$ classes were selected to constitute SceneParse150 challenge. It consists of $20,210$ training images, $2,000$  validation images, and covers $150$ fine-grained labeled classes. Models are trained for $160$k iterations, with a batch size of $16$ and a crop size of $640 \times 640$ pixels when using Swin-B and Swin-L as backbone; otherwise a crop size of $512 \times 512$ is used.

\textbf{COCO-Stuff-10K} \cite{caesar2018cocostuff} is a subset of the COCO dataset \cite{lin2015microsoft} for semantic segmentation. It consists of $9$k images for training and $1$k images for testing, covering $171$ semantic-level categories. For training, all SenFormer models were trained for $80$k iterations, with a batch size of $16$ and a crop size of $512 \times 512$.

\textbf{Pascal Context} \cite{pascal} training set contains  $4,996$ images covering $59$ classes and the testing set contains  $5,104$ images. The data come from the PASCAL VOC 2010 contest \cite{Everingham2009ThePV}, where annotations for the whole scene have been added. All SenFormer models were trained for $40$k iterations, with a batch size of $16$ and a crop size of $480 \times 480$.

\textbf{Cityscapes} \cite{cordts2016cityscapes} is a high-resolution dataset of $5,000$ street-view images with $19$ semantic classes. Conventionally, the dataset is split into a training set of $2,975$ images and a validation set of $500$ images. All SenFormer models were trained for $100$k iterations, with a batch size of $8$ and a crop size of $512 \times 1024$.


\section{Detailed Learners Architecture}\label{sec:learner_arch}
Hereafter, we detail the architecture of the learners.
Each decoder is a transformer composed of $L$ layers.
In a nutshell, a single Transformer Decoder block consists of three successive operations: \textit{Cross-Attention} (CA), \textit{Self-Attention} and \textit{Multi-Layer Perceptron} layers.

First, the multi-scales features $\{P_2,P_3,P_4,P_5\}$ are reshaped into a set of tokens $\{z_2,z_3,z_4,z_5\}$ where $z_i \in \mathcal{R}^{n_i \times d}$, $n_i= \frac{HW}{2^i}$ is the number of token and $d$ is the numbers of channels in all the feature maps of the FPN.

In the CA layer, the FPN's token features $z_i$'s are linearly transformed through matrix multiplication to acts as the \textit{keys} and \textit{values}, and the class embedding $\textbf{cls}_i$'s as the \textit{queries}. For $i \in \{2, 3, 4, 5\}$ the CA operation is as follows:
\begin{align}
    &K = z_i W_K, V = z_i W_V, Q =  LN(\textbf{cls}_i) W_Q, \\ \nonumber
    &\text{where } W_K, W_V, W_Q \in \mathcal{R}^{D \times D} \\ \nonumber
    &CA(Q_i, P_i) = \textbf{cls}_i + softmax(\frac{QK^T} {\sqrt{D}})V
\end{align}
where \textit{softmax} denotes the softmax function applied along the last dimension and \textit{LN} denotes the "Layer Normalization" operation.
This operation aims at distilling the knowledge gained by the backbone contained in the features $P_i$ into the class embeddings $\textbf{cls}_i$. This way, the model will retain the information of \emph{what is it for a feature token $z_i$ to represent a specific class.} Indeed, since the class embedding vector is used to get the final prediction via a dot product with the features tokens, during the back-propagation the class embedding tensor $cls_i^k$ (that represents the $k^{th}$ class) will be encouraged to become similar to the tokens of $z_i$ that correspond to the $k^{th}$ class, according to the ground truth segmentation maps.

Then, the \textit{Self-Attention} layer enables sharing the information acquired during the Cross-Attention across the class embedding vectors. Eventually, an \textit{Multi-Layer Perceptron} layer is used to propagate the information across the channel dimensions. 

Overall, each decoder is composed of $L$ layers of decoder blocks, and its prediction is obtained via a dot product between the class embeddings $\textbf{cls}_i$ and the corresponding feature pyramid feature $P_i$.

\subsection{Feature Pyramid Network}
The set of features extracted by the backbone $\{ C_2, C_3, C_4, C_5 \}$, is enhanced by the FPN \cite{lin2017feature} to obtain a feature pyramid that has strong spatial and semantics at all scales. To do so, $\{ C_2, C_3, C_4, C_5 \}$ undergo a linear projection to set the channel dimension of each scale to a fixed size denoted as $d$(fixed to $512$). Consecutive levels are then upsampled to the same size and merged by element-wise addition. Eventually, the merged features are processed by a $3 \times 3$ convolution to alleviate the aliasing effect of the upsampling.
In sum, the output set of features of the FPN $\{ P_2, P_3, P_4, P_5 \}$ is obtained by:
\begin{align}
    P_i^{\prime}  &= \textit{Conv}_{1 \times 1}\left( C_{i}\right), \quad i \in \{2,3,4,5\} \\ \nonumber
    P_5 &= P_5^{\prime} \\ \nonumber
    P_i &= \textit{Conv}_{3\times 3} \left( P_{i}^{\prime} + \textit{Upsample}(P_{i+1}^{\prime})\right),  \quad i \in \{2,3,4\} \label{eq:FPN}
\end{align}
with $\textit{Conv}_{ \cdot \times \cdot}$ being a convolution with $\cdot \times \cdot$ kernel and \textit{Upsample} being the nearest neighbor upsampling operation.
\subsubsection{Feature Pyramid Network Transformer}
We empirically found that while introducing marginal changes to the implementation and a minimal computational cost, replacing the $3\times 3$ convolution by a \textit{transformer block} in the FPN increases the segmentation performance for SenFormer, see Table \ref{tab:ablation}.a.
In practice, we use the window-based transformer block (denoted as WTB) of \cite{liu2021swin} to limit the memory footprint overhead. We name this enhanced FPN version as Feature Pyramid Network Transformer-enhanced (FPNT), where
\begin{align}
    P_i^{\prime}  &= \textit{Conv}_{1 \times 1}\left( C_{i}\right), \quad i \in \{2,3,4,5\} \\ \nonumber
    P_5 &= P_5^{\prime} \\ \nonumber
    P_i &= \textit{WTB}\left( P_{i}^{\prime} + \textit{Upsample}(P_{i+1}^{\prime})\right),  \quad i \in \{2,3,4\} \label{eq:FPNT}
\end{align}

\begin{table} [h!]
\setcounter{table}{12}
\caption{Ablation studies related to SenFormer architectural choices. Models are trained on ADE20K validation for 100k iterations with a ResNet-50 backbone pretrained on ImageNet-1K \cite{russakovsky2015imagenet}}\label{tab:ablation}

\centering
\qquad
\subfloat[FPN vs FPNT.]{\begin{tabular}{cc}
\toprule
method & \textit{mIoU} \\
 \midrule[\heavyrulewidth]

none & 38.15  \\
FPN & 42.7   \\
\rowcolor{almond} FPNT & \textbf{43.6}  \\
\bottomrule
\end{tabular}}
\qquad 
\subfloat[Normalization strategy.]{\begin{tabular}{cc}
\toprule
\multirow{2}{*}{method}  &\multirow{2}{*}{\textit{mIoU}}   \\
\\
 \midrule[\heavyrulewidth]

Post-Norm & 42.63  \\
\rowcolor{almond} Pre-Norm & \textbf{43.6}  \\
\bottomrule
\end{tabular}}

\end{table}

\section{Additional Experiments}

\begin{table*}[]
    \centering
    \caption{Comparison of the "repeated" sharing strategy vs no weight sharing (none) on multiple benchmark datasets. }\label{tab:Repeated}
    \begin{tabular}{c c c c c c c}
    \toprule
         backbone & sharing & params. & ADE20k & Pascal & COCO & Cityscapes \\
          \midrule[\heavyrulewidth]
        \multirow{2}{*}[-0.1em]{ResNet-50} &  none &  144M &\textbf{44.6} & \textbf{53.2} & 39.0 & \textbf{78.8}\\
         &  \cellcolor{almond} repeated   & \cellcolor{almond}55M & \cellcolor{almond}44.3 & \cellcolor{almond}\textbf{53.2} & \cellcolor{almond}\textbf{40.0} & \cellcolor{almond}\textbf{78.8}\\
        \hline 
        \multirow{2}{*}[-0.1em]{ResNet-101} &  none &  163M & 46.5 & \textbf{55.1} & 39.6 & \textbf{80.3} \\
        &  \cellcolor{almond}repeated &  \cellcolor{almond}79M & \cellcolor{almond}\textbf{46.9} & \cellcolor{almond}54.6 & \cellcolor{almond}\textbf{41.0} & \cellcolor{almond}79.9 \\
        \midrule
        \multirow{2}{*}[-0.1em]{Swin-L} &  none &  314M & \textbf{53.1} & 62.4 & 49.1 & 82.2 \\
        &  \cellcolor{almond}repeated &  \cellcolor{almond}233M & \cellcolor{almond}\textbf{53.1} & \cellcolor{almond}\textbf{63.1} & \cellcolor{almond}\textbf{49.8} & \cellcolor{almond}\textbf{82.8} \\
        \bottomrule
    \end{tabular}
\end{table*}

\textbf{Weight sharing.} In Table \ref{tab:Repeated} we compare on multiple datasets SenFormer performance when no weight sharing is used (\textit{none}) and the default weight sharing setting, where the same decoder block is recursively used $L$ times (\textit{repeated}). Despite having significantly fewer parameters, our weight sharing strategy has similar performance to when no sharing is used.\\

\textbf{Multi-scales feature generation.} In Table \ref{tab:ablation}.a we demonstrate the benefit of the feature pyramid network transformer (FPNT) over the traditional FPN for SenFormer. We observe that FPNT is $0.9$ \textit{mIoU} better than the FPN baseline (with ResNet50 as backbone). Also, not using any FPN-like method significantly reduces the performance, which can be explained by the fact that each learner must receive semantically and spatially strong features. 

\section{Additional discussions}\label{sec:add-discusions}
This section further discusses the impact of the learners' architectural choices on their predictions. In particular, we study the behavior of the class embeddings after convergence. Furthermore, we show how the "pre-norm" strategy may help the class embeddings act as a memory bank.\\

\textbf{Class embeddings acting like a memory bank.} In SenFormer, for a given learner, each class embedding vector represents a unique class and is used to produce the learner's prediction for that given class. Hence, we expect the class embedding to retain specific information about that class. Accordingly, at the end of the training, the different learners should converge to different vectors (as they represent different classes).
 
\begin{figure}[] 
    \centering
    
    \includegraphics[width=1\linewidth, trim = 0.8cm 0cm 1.2cm 1.4cm, clip]{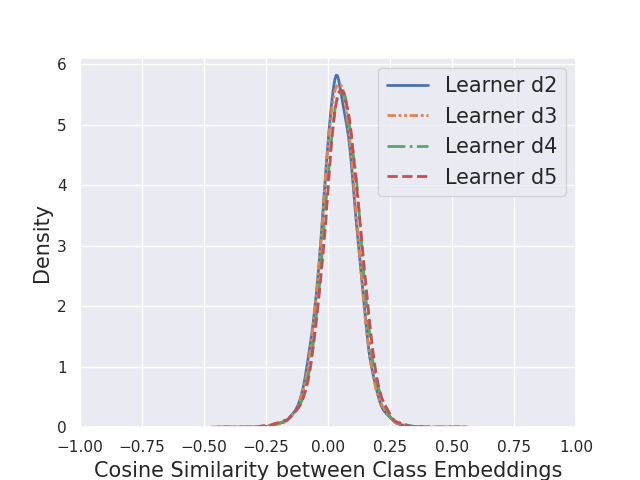}
    \caption{Distribution of the cosine similarity between the different class embeddings for each learner in SenFormer.} \label{fig:density}
\end{figure}

 Figure \ref{fig:density} shows the distribution density of the cosine similarity between the different class embeddings of SenFormer trained on ADE20K with ResNet-50 as backbone; \textit{i.e.} for the $i$-th learner the distribution of the following set $\left\{\cos(cls_i^k, cls_i^l), \; (k,l)\in\{1,..,N_c-1\}\times \{k,.., N_c\} \right\} $. 
 
 As expected, although ADE20K has a large number of classes, the density curves are close to the origin, suggesting that the different class embeddings converge to different vectors. Hence, for a given learner $i$, the class embeddings $\textbf{cls}_i = [cls^1_i, \dots, cls^{N_c}_i]$ of SenFormer effectively acts as a \emph{memory bank} that is used by the decoder to assess how likely a given feature token of $P_i$ represents a certain class.



\textbf{Pre-Norm vs Post-Norm.} Early versions of transformers \cite{vaswani2017attention} as well as recent applications to object detection and segmentation \cite{carion2020endtoend,cheng2021perpixel} applied layer normalization after the skip connection (post-norm), while recent implementations tend toward using pre-normalization setting. Table \ref{tab:ablation}.b shows that a pre-normalization strategy performs best for our architecture. We believe that, by leaving the skip connection pathway unaltered, the pre-norm setting ease the information flow from the ground truth supervision to the input class embeddings of the learner, therefore fostering the \emph{"memory bank"} mechanism described above. It may also explain why \cite{carion2020endtoend,cheng2021perpixel} do not benefit from the use of pre-norm. Indeed, in DETR and MaskFormer, each \textit{query embedding} vector (the equivalent of our class embedding) does not correspond to a unique class, but is rather dynamically routed to a class by using a bipartite matching, therefore the learned embeddings do not act as a "memory bank".

\section{Visualization}
\begin{figure*}[] 
    \centering
    
    \includegraphics[width=0.9\linewidth]{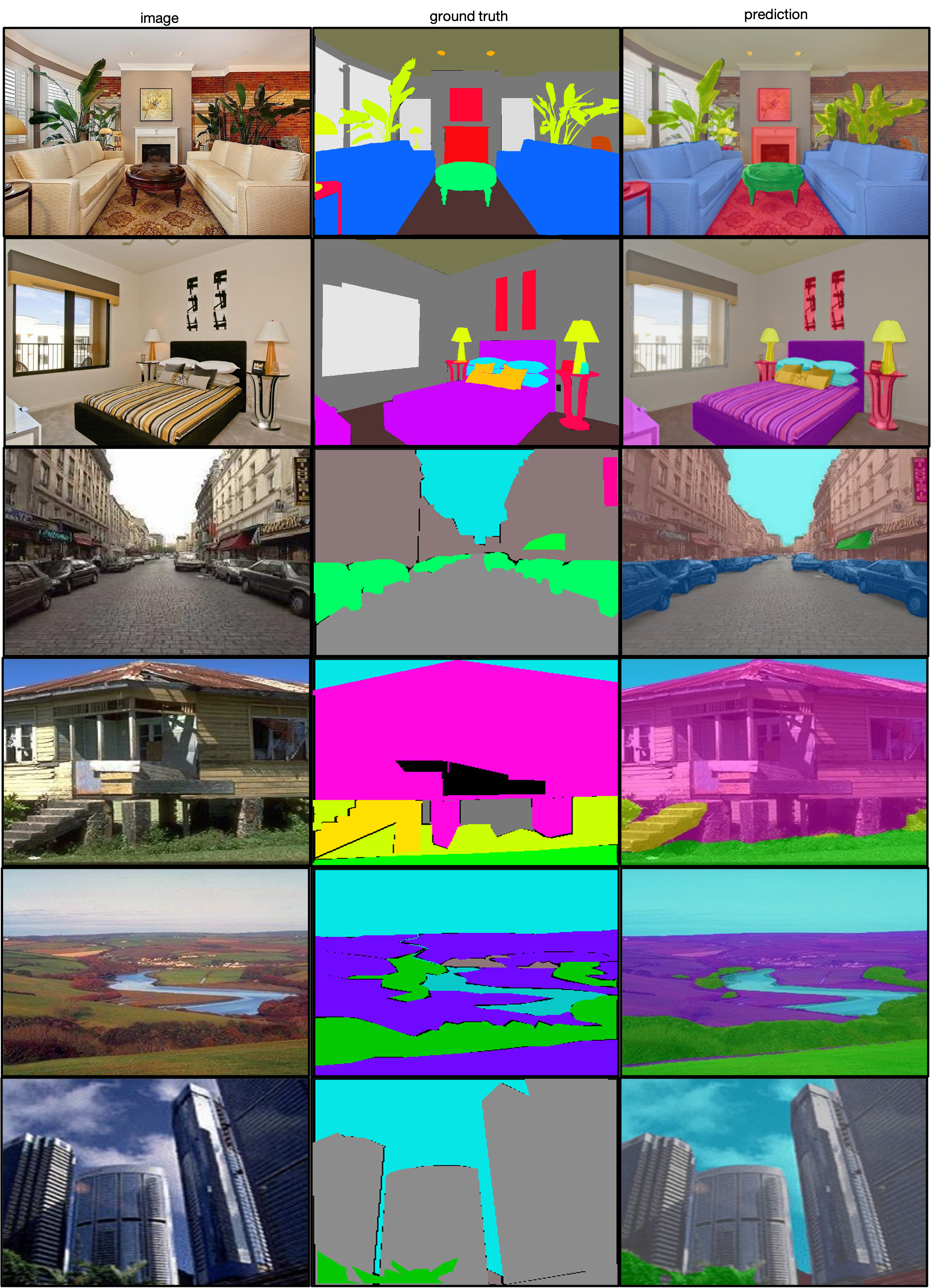}
    \caption{Visualization of SenFormer segmentation predictions on ADE20K validation with Swin-L backbone.} \label{fig:PredSamples}
\end{figure*}

Figure \ref{fig:PredSamples}, we visualize sample segmentation predictions of SenFormer with Swin-L backbone on ADE20K validation.  

\end{document}